%% file: main.tex
\documentclass[runningheads]{llncs}

\usepackage{eccv}

\usepackage{eccvabbrv}

\usepackage{amsmath}
\usepackage{amssymb}
\usepackage{booktabs,multirow,xcolor,colortbl,graphicx}
\usepackage[accsupp]{axessibility}  

\usepackage{hyperref}

\usepackage{orcidlink}

\begin{document}

\title{Not All Redundant Tokens Are Alike: Analyzing Visual Token Pruning through Token Roles} 

\titlerunning{Not All Redundant Tokens Are Alike}

\author{Hyeonyu Kim\inst{1}\orcidlink{0009-0007-3079-0647} \and
Sehwan Lim\inst{1}\textsuperscript{*}\orcidlink{0009-0006-5943-3902} \and
Youngwon Choi\inst{1}\textsuperscript{*}\orcidlink{0009-0005-2528-9263} \and\\
Taeyoun Kwon\inst{1}\textsuperscript{*}\orcidlink{0009-0005-2875-2612} \and
Jaejin Kim\inst{1}\textsuperscript{*}\orcidlink{0009-0001-4499-3038}}

\authorrunning{Kim et al.}

\institute{Maum AI Inc., Republic of Korea\\
\email{\{hykim, sehwan.lim, youngwonchoi, taeyoun.kwon, jaejin.kim\}@maum.ai}
}

\maketitle
\begingroup
\renewcommand{\thefootnote}{*}
\footnotetext{These authors contributed equally. Their ordering was determined by random draw.}
\endgroup

\input{sections/0_abstract}    
\input{sections/1_intro}
\input{sections/2_related}
\input{sections/3_method}
\input{sections/5_conc}

\section{Acknowledgments}

This research was supported by Culture, Sports and Tourism R\&D Program through the Korea Creative Content Agency grant funded by the Ministry of Culture, Sports and Tourism in 2026 (Project Name: Development of AI-based personalized cultural and arts learning services using smart device, Project Number: RS-2026-25524629). 

%
%
\clearpage
\bibliographystyle{splncs04}
\bibliography{main}
\end{document}

%% file: sections/0_abstract.tex
\begin{abstract}
  Vision-language models (VLMs) process an image as a sequence of visual tokens, which creates a substantial computational bottleneck during inference. 
  Recent visual token pruning methods address this issue by removing seemingly redundant tokens, yet it remains unclear how these pruning decisions relate to the functional roles of visual tokens. 
  In this work, we analyze visual token pruning through the lens of token roles identified by EmbedLens. 
  We first show that representative pruning methods exhibit distinct token-role biases, but these biases do not directly correlate with downstream performance. 
  To better understand this behavior, we refine the token-role assignment procedure and evaluate role-protected pruning variants. 
  Our results show that preserving non-alive tokens can sometimes maintain or improve performance, suggesting that tokens with weak direct semantic alignment may still affect model behavior under pruning.
  Our code is publicly available at \url{https://github.com/jaykim9870/Not_All_Redundant_Tokens_Are_Alike}.
  \keywords{Visual Token Pruning \and Vision Language Models \and EmbedLens}
\end{abstract}

%% file: sections/1_intro.tex
\section{Introduction}

Vision language models (VLMs) achieve strong multimodal reasoning by converting an image into a sequence of visual tokens and feeding them into an LLM~\cite{liu_visual_2023, liu2024improved}. 
However, this design also introduces a major efficiency bottleneck: visual tokens often dominate the input sequence length, increasing prefill latency, memory consumption, and KV-cache cost. 
To address this issue, recent visual token pruning methods reduce the number of visual tokens while attempting to preserve downstream performance~\cite{shang_llava-prumerge_2025, yang_visionzip_2025, li_tokenpacker_2025, zhang_sparsevlm_2025, leonardis_image_2025, wen_stop_2025, alvar_divprune_2025}. 
These methods typically select tokens according to criteria such as attention, feature similarity, duplication, text relevance, or diversity.

Despite their success, it remains unclear what kinds of visual tokens are removed by these pruning methods. 
Recent interpretability studies suggest that visual tokens in VLMs are not homogeneous carriers of image semantics. 
In particular, EmbedLens analyzes projected visual tokens in the LLM input embedding space and reveals a sparse token structure consisting of alive tokens, sink tokens, and dead tokens~\cite{fan2026visual}. 
Alive tokens are more closely associated with image-specific semantics, whereas sink and dead tokens are often repetitive, image-agnostic, or weakly aligned with lexical semantics. 
This observation naturally suggests a simple expectation: visual token pruning should preferentially remove non-alive tokens, while preserving alive tokens that carry image-specific information.

In this work, we examine whether this expectation holds for existing visual token pruning methods. 
We connect EmbedLens-style token role analysis with representative training-free pruning methods, including FastV~\cite{leonardis_image_2025}, DART~\cite{wen_stop_2025}, and DivPrune~\cite{alvar_divprune_2025}. 
We first analyze which token roles are pruned under different token budgets across ten VLM benchmarks. 
Our results show that pruning methods exhibit distinct token-role biases: for example, DivPrune tends to prune more non-alive tokens, while FastV and DART often retain sink tokens. 
However, these token-role distributions alone do not fully explain downstream performance. 
Methods that remove more non-alive tokens do not always achieve better performance, suggesting that the utility of a token cannot be determined solely from its token role.

To study this more directly, we refine token role assignment in EmbedLens analysis. 
The original EmbedLens procedure assigns visual token roles through cluster-level textual reference token IDs in the LLM embedding space. 
We observe that this text-ID-based assignment can misclassify a substantial portion of sink and alive tokens, partly because centroid-level textual proximity does not always transfer to token-level role assignment. 
We therefore introduce a centroid-based assignment strategy with an additional alive-token centroid, which provides role assignments that are more consistent with the cluster-derived assignments while requiring only a small set of global centroids. 
This enables us to analyze token roles consistently across pruning methods, datasets, and attention patterns.

Using this refined role assignment, we then probe the functional contribution of each token role under pruning. 
Instead of simply removing a token role from the input, we evaluate role-protected pruning strategies in which tokens of a specified role are excluded from the pruning candidate set. 
This allows us to ask whether preserving a particular role changes downstream performance. As expected, preserving alive tokens is generally beneficial. 
More interestingly, preserving non-alive tokens, including dead tokens, does not necessarily harm performance and can sometimes improve it. 
This suggests that tokens with weak individual semantic alignment may still play a role under pruning, especially when multiple token categories are jointly removed.

Finally, we analyze cross-role attention patterns to better understand this behavior. 
We measure group-level attention mass and pairwise attention scores among token roles under controlled token removal and DivPrune-based pruning. 
Our analysis indicates that dead tokens have low per-token attention but can occupy a non-negligible group-level attention budget due to their large group size. 
Under DivPrune, protecting dead tokens partially restores their interaction patterns while redistributing attention away from other token groups. 
These observations suggest that, although dead tokens may not serve as direct carriers of visual semantics, their aggregate or structural interactions can still influence model behavior under token pruning.


%% file: sections/2_related.tex
\section{Related Work}

\subsubsection{Multimodal Alignment and Interpretability in VLMs}

Modern VLMs typically connect a pretrained vision encoder to an LLM through a modality adapter, making visual--textual alignment a central factor for multimodal reasoning~\cite{liu_visual_2023, liu2024improved}. 
SEA explicitly studies token-level visual--textual embedding alignment and shows that insufficient alignment can limit both performance and interpretability~\cite{yin2025sea}. 
More closely related to our work, EmbedLens analyzes projected visual tokens by directly comparing them with the LLM input embedding space, revealing that visual tokens form distinct functional categories, including dead tokens, sink tokens, and alive tokens~\cite{fan2026visual}. 
LatentLens further supports this embedding-space perspective by showing that visual tokens can often be interpreted through nearby contextualized textual representations, and that conventional LogitLens-style decoding may underestimate their interpretability~\cite{krojer2026latentlens}.

This line of work is also connected to broader representation-level interpretability methods. 
The Tuned Lens formalizes and improves the earlier Logit Lens approach by learning layer-wise affine transformations, providing a more faithful view of latent predictions in autoregressive language models~\cite{belrose2023eliciting}. 
Beyond vocabulary-based probing, recent studies suggest that semantic or behavioral properties can be localized in representation space. 
The Platonic Representation Hypothesis argues that representations across models and modalities may converge toward shared statistical structures, while representation engineering and activation steering studies show that specific directions in activation space can be used to monitor or control model behavior~\cite{huh2024platonic, zou2023representation, li2023inference}. 
In VLMs, Visual Attention Sink further demonstrates that tokens receiving high attention are not necessarily semantically meaningful visual evidence, indicating that attention-based importance and functional token roles can diverge~\cite{kang2025see}. 

While these interpretability studies mainly aim to understand VLM representations, visual token pruning studies primarily optimize the accuracy--efficiency trade-off by selecting which visual tokens to retain or discard. 
As a result, it remains unclear how the token categories revealed by embedding-space analyses interact with actual pruning decisions. 
In this work, we bridge these two lines of research by connecting EmbedLens-style token role analysis with representative visual token pruning methods.

\subsubsection{Visual Token Redundancy in VLMs}

Although modern VLMs achieve strong vision-language understanding by integrating image-derived visual tokens into an LLM, these visual tokens often dominate the input sequence length, especially for high-resolution images or video inputs, resulting in a major bottleneck for real-time applications and edge deployment~\cite{shang_llava-prumerge_2025, chu_mobilevlm_2023, li_tokenpacker_2025, bolya_token_2023, lin_boosting_2025, zhang_sparsevlm_2025}. 
At the same time, visual inputs often contain substantial spatial and temporal redundancy, and prior studies have observed that not all visual tokens contribute equally to response generation~\cite{yang_visionzip_2025}, which motivates a range of visual token pruning and compression methods. 

One line of work uses attention as a proxy for token importance. 
FastV prunes low-attention visual tokens after shallow LLM layers, based on the observation that many visual tokens receive sharply reduced attention as the model processes inputs~\cite{leonardis_image_2025}. 
PDrop progressively removes visual tokens across layers, reflecting the view that visual redundancy becomes more pronounced in deeper layers~\cite{xing_pyramiddrop_2025}. 

Another line of work performs pruning or merging in the vision encoder space. 
PruMerge and VisionZip exploit attention distributions and feature similarity within the visual encoder to select informative tokens, and then aggregate the remaining redundant tokens---either merging them into the retained tokens (PruMerge) or condensing them into a small set of additional contextual tokens (VisionZip)~\cite{shang_llava-prumerge_2025, yang_visionzip_2025}. 

To further incorporate task relevance, text-guided methods often condition token selection on the textual instruction. 
SparseVLM, for example, performs pruning based on query-relevant text tokens, making the retained visual tokens more closely aligned with the current instruction~\cite{zhang_sparsevlm_2025}. 
More recent methods reconsider the selection criterion itself. DART argues that reducing duplication among visual tokens can be more important than estimating individual token importance, while DivPrune formulates pruning as a diversity maximization problem over the retained subset~\cite{wen_stop_2025, alvar_divprune_2025}. 

Despite this progress, it remains underexplored whether tokens that appear redundant or semantically weak in embedding-space analyses are actually dispensable during pruning, or whether they still contribute through interactions with other tokens.



%% file: sections/3_method.tex
\section{Our Approach}

\subsection{Preliminaries}\label{sec:method:preliminaries}

\subsubsection{EmbedLens}

Vision language models (VLMs) typically encode an image using a vision encoder, and then map a sequence of encoded features into the language model's input space through a projector. 
Let $\mathbf{v} \in \mathbb{R}^d$ denote the projected visual token, and let $\mathbf{W}_E \in \mathbb{R}^{|\mathcal{V}| \times d}$ be the input embedding matrix of the language model, where $\mathbf{e}_i$ denotes the embedding of vocabulary token $i$.

Unlike LogitLens-style analyses, which decode hidden states through the language modeling head, EmbedLens directly compares a target representation with $\mathbf{W}_E$. 
This makes it particularly suitable for examining whether a projected visual token is already located near meaningful lexical concepts before substantial transformation inside the LLM.

Formally, given a target representation $\mathbf{h}$ to analyze, EmbedLens retrieves the nearest vocabulary tokens according to cosine similarity:
\begin{equation}
\mathrm{EmbedLens}_k(\mathbf{h})
=
\mathrm{TopK}_{i \in \mathcal{V}}
\left(
\frac{\mathbf{h}^{\top}\mathbf{e}_i}
{\|\mathbf{h}\|_2 \|\mathbf{e}_i\|_2}
\right).
\end{equation}
For a projected visual token $\mathbf{v}$, the retrieved vocabulary tokens provide a discrete textual proxy for its location in the LLM input space. 
The same operation can also be applied to a cluster centroid, allowing a visual-token cluster to be assigned a representative textual anchor.

Using this probing tool, Fan et al.~\cite{fan2026visual} observe that projected visual tokens form a sparse structure consisting of three distinct groups: sink tokens, dead tokens, and alive tokens. 
They first cluster normalized projected visual tokens using an anchor-based criterion. 
For an anchor token $\mathbf{v}_a$, its cluster is defined as
\begin{equation}
C_a =
\left\{
j \;:\;
\frac{\mathbf{v}_a^{\top}\mathbf{v}_j}
{\|\mathbf{v}_a\|_2 \|\mathbf{v}_j\|_2}
\ge \tau
\right\},
\end{equation}
where $\tau=0.9$. 

For each image, the largest cluster $C_0$ contains a large fraction of visual tokens and remains highly homogeneous across images. 
These tokens are far from the text embedding manifold and do not yield coherent lexical semantics under EmbedLens. 
The paper therefore identifies this dominant, repetitive, and text-distancing cluster as \emph{dead tokens}.

The paper further distinguishes two types of visual sink tokens. 
The first are \emph{ViT sink tokens}, which are inherited from the vision encoder and exhibit large activation norms in the final layer of CLIP. 
Specifically, a visual token is considered a ViT sink token when the $L_2$ norm of its final-layer CLIP hidden state exceeds 75. 
The centroid of these tokens is nearly identical across images.
The second are \emph{LLM sink tokens}, which emerge inside the language backbone. 
Following prior work~\cite{kang2025see}, EmbedLens defines a sink score based on a small set of sink dimensions $D_{\mathrm{sink}}$. 
For LLaMA-2-7B-based models, $D_{\mathrm{sink}}=\{1415,2533\}$. Given a hidden state $\mathbf{x} \in \mathbb{R}^D$, the score is defined as
\begin{equation}
\operatorname{Sink}(\mathbf{x})
=
\max_{d \in D_{\mathrm{sink}}}
\frac{\mathbf{x}[d]}
{\sqrt{\frac{1}{D}\sum_{r=1}^{D}\mathbf{x}[r]^2}}.
\end{equation}
A visual token is treated as an LLM sink when $Sink(\mathbf{x}) \ge 20$. 
Their centroid is also stable across images.

After removing dead tokens and sink tokens, the remaining visual tokens are defined as \emph{alive tokens}. 
These tokens lie closer to the text-semantic region and serve as the primary carriers of image-specific information. 


\subsubsection{Visual Token Pruning}
We consider three representative training-free visual token pruning methods in our work, namely FastV~\cite{leonardis_image_2025}, DART~\cite{wen_stop_2025}, and DivPrune~\cite{alvar_divprune_2025}. 
They adopt distinct selection criteria: attention-based importance, token duplication, and global diversity, respectively.

\textbf{FastV}~\cite{leonardis_image_2025} observes that visual tokens receive substantially lower attention than textual tokens in the deep layers of VLMs. 
Based on this observation, it treats the attention score that a token \emph{receives} as an indicator of its importance. 
Let $\mathbf{A}^{(K)}$ denote the head-averaged self-attention matrix at the $K$-th LLM layer, where $\mathbf{A}^{(K)}_{j,i}$ is the attention weight from query token $j$ to key token $i$. 
The importance of a visual token $\mathbf{x}_i$ is defined as the average attention it receives from the subsequent tokens in the sequence,
\begin{equation}
\phi_{\mathrm{attn}}(\mathbf{x}_i)
= \frac{1}{|\mathcal{Q}_i|} \sum_{j \in \mathcal{Q}_i}
\mathbf{A}^{(K)}_{j,i},
\label{eq:fastv_score}
\end{equation}
where $\mathcal{Q}_i$ denotes the set of tokens that attend to $\mathbf{x}_i$ under the causal mask, including both visual and textual tokens. Visual tokens with the lowest scores in Eq.~\eqref{eq:fastv_score} are discarded from all subsequent layers.

\textbf{DART}~\cite{wen_stop_2025} identifies several limitations of attention-based approaches and shifts the selection criterion from importance to duplication. 
It selects a small set of pivot tokens $\mathcal{P} = \{\mathbf{p}_1, \dots, \mathbf{p}_k\}$, according to criteria such as the K-norm. 
The duplication score between a pivot $\mathbf{p}_i$ and a visual token $\mathbf{x}_j$ is defined as

\begin{equation}
\mathrm{dup}(\mathbf{p}_i, \mathbf{x}_j)
=
\frac{\mathbf{p}_i^{\top}\mathbf{x}_j}
{\|\mathbf{p}_i\|_2 \, \|\mathbf{x}_j\|_2}.
\label{eq:dart_dup}
\end{equation}
Only tokens whose scores do not exceed the threshold are retained. 
DART requires no access to attention maps and is fully compatible with FlashAttention.
Moreover, the resulting output deviation is theoretically bounded via the Hausdorff distance.



\textbf{DivPrune}~\cite{alvar_divprune_2025} further observes that pruning guided by a small number of reference tokens remains local and the retained subset may still fail to represent the original tokens at high reduction ratios. 
It instead directly maximizes the diversity of the entire retained subset. 
Token pruning is formulated as a Max--Min Diversity Problem (MMDP), which selects from $M$ visual tokens $\mathbf{E}_v$ a subset of size $\tilde{M}$ maximizing the minimum pairwise distance,

\begin{equation}
\tilde{\mathbf{E}}_v
=
\arg\max
\Bigl[
\min_{\boldsymbol{\gamma}, \boldsymbol{\omega} \in S}
d(\boldsymbol{\gamma}, \boldsymbol{\omega})
\, \colon \,
\forall S \subset \mathbf{E}_v,\;
|S| = \tilde{M}
\Bigr],
\label{eq:divprune_mmdp}
\end{equation}
where the distance $d(\cdot,\cdot)$ is measured by the cosine distance. 
Eq.~\eqref{eq:divprune_mmdp} is solved with a greedy algorithm. 
Specifically, the first token is chosen by pairwise distances, and each subsequent token is the one whose minimum distance to the selected set is maximal. 



\begin{figure}[t]
    \centering
    \includegraphics[width=\linewidth]{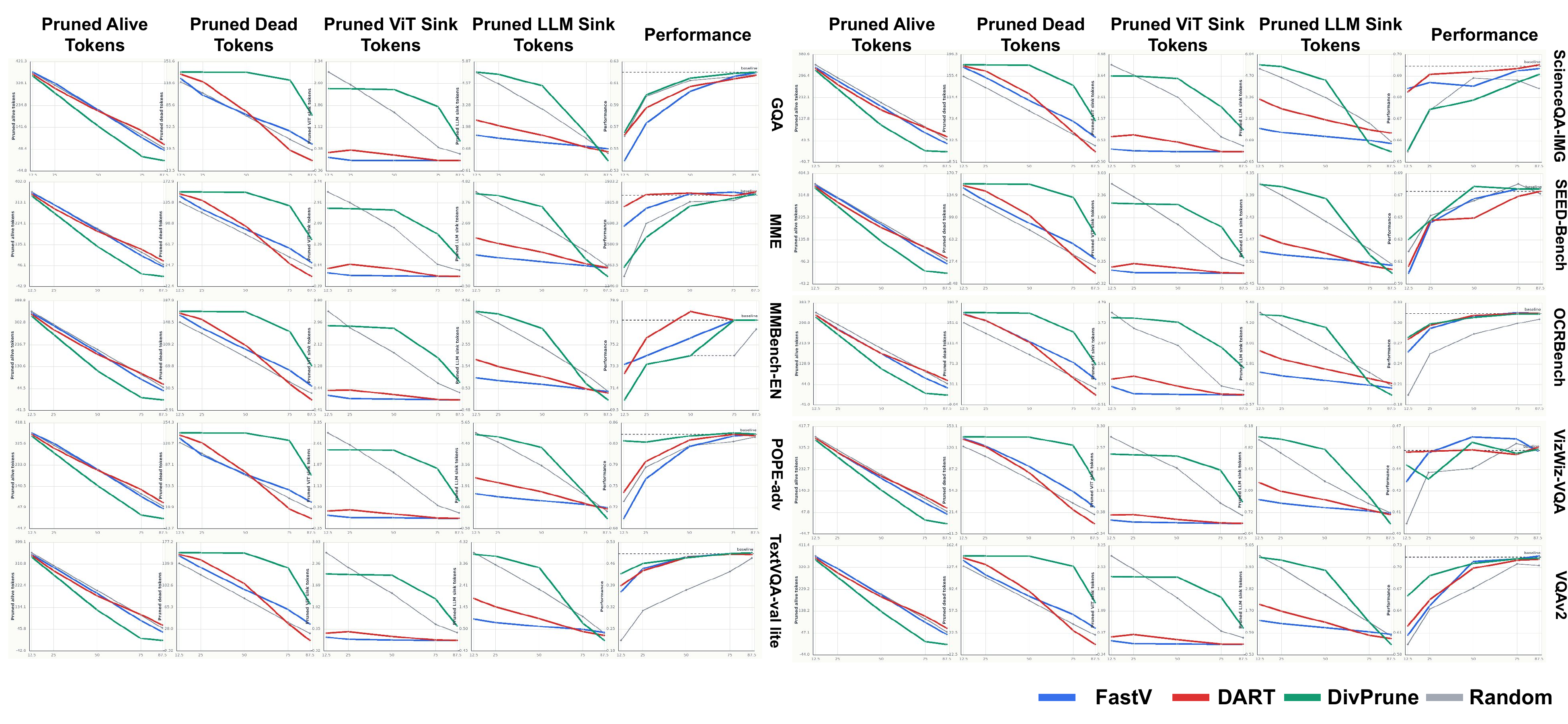}
    \caption{
    Token-category distribution and downstream performance under different token budgets. For FastV, DART, and DivPrune, we report the composition of pruned tokens over four token categories and the corresponding performance across ten benchmarks. The x-axis denotes the token budget, while the y-axis indicates either the number of pruned tokens in each category or the downstream task performance.
    }
    \label{fig:token_dist_original}
\end{figure}

\subsection{Token-Role Biases in Existing Pruning Methods}\label{sec:method:observation}

As introduced in Section~\ref{sec:method:preliminaries}, EmbedLens \cite{fan2026visual} reveals that projected visual tokens $\mathbf{v}$ form a sparse and redundant structure in the LLM input space. 
Specifically, visual tokens can be categorized into four groups: dead tokens, ViT sink tokens, LLM sink tokens, and alive tokens, and removing non-alive tokens does not degrade performance and can even improve it in some cases.

Building on these token roles, we investigate which types of visual tokens are pruned by existing visual token pruning methods under different token budgets. 
In particular, we apply FastV~\cite{leonardis_image_2025}, DART~\cite{wen_stop_2025}, and DivPrune~\cite{alvar_divprune_2025} to LLaVA-v1.5-7B~\cite{liu2024improved} and analyze the distribution of the removed tokens in terms of the four token categories. 
We report results under five token budgets: 12.5\%, 25\%, 50\%, 75\%, and 87.5\%.

For evaluation, we use ten representative VLM benchmarks: GQA~\cite{hudson2019gqa} (testdev), MME~\cite{fu2026mme}, MMBench~\cite{liu2024mmbench} (English dev, lite subset), POPE~\cite{li2023pope} (adversarial split), TextVQA~\cite{singh2019textvqa} (validation, lite subset), ScienceQA-IMG~\cite{lu2022scienceqa}, SEED-Bench~\cite{li2023seedbench} (lite subset), OCRBench~\cite{liu2024ocrbench}, VizWiz-VQA~\cite{gurari2018vizwiz} (validation, lite subset), and VQAv2~\cite{goyal2017vqav2} (validation, lite subset).
For each benchmark, we report the primary metric used by the corresponding evaluation task: exact-match accuracy for GQA, overall multiple-choice accuracy for MMBench, the sum of perception and cognition scores for MME, the normalized final score for OCRBench, F1 score on the adversarial split for POPE, multiple-choice accuracy for ScienceQA-IMG, image-subset multiple-choice accuracy for SEED-Bench, and VQA-style soft accuracy for TextVQA, VizWiz-VQA, and VQAv2.

Fig.~\ref{fig:token_dist_original} presents the token-category distribution of the pruned tokens together with the corresponding performance on ten benchmarks. 
Overall, we observe that the three pruning methods do not consistently favor certain token categories. Instead, each method exhibits a distinct token-role bias.
Specifically, DivPrune tends to select a larger proportion of non-alive tokens, whereas FastV and DART are less likely to prune sink tokens. 
This behavior can be understood from their pruning mechanisms: FastV relies on attention scores, while DART selects pivot tokens based on the K-norm. As a result, ViT sink tokens, which have large norms in the last vision encoder layers, and LLM sink tokens, which tend to receive high attention within the LLM, are more likely to be retained. 
Although EmbedLens suggests that ViT sink tokens are often disregarded within the LLM backbone despite their large norms, our results suggest that these tokens still receive high attention in the early LLM layers.

Interestingly, this tendency remains consistent across different benchmarks. Although FastV incorporates text instructions into its token-pruning criterion, the token-role bias remains consistent across different tasks. 
However, such token-role bias does not necessarily translate into downstream task performance. 
For example, DivPrune, which prunes a larger proportion of non-alive tokens, achieves the best performance on several benchmarks, whereas FastV outperforms the other methods on some tasks. 
In the following sections, we analyze why this phenomenon occurs and how the distinct categories of visual tokens interact with visual token pruning methods and downstream task performance.

\begin{figure}[t]
    \centering
    \includegraphics[width=\linewidth]{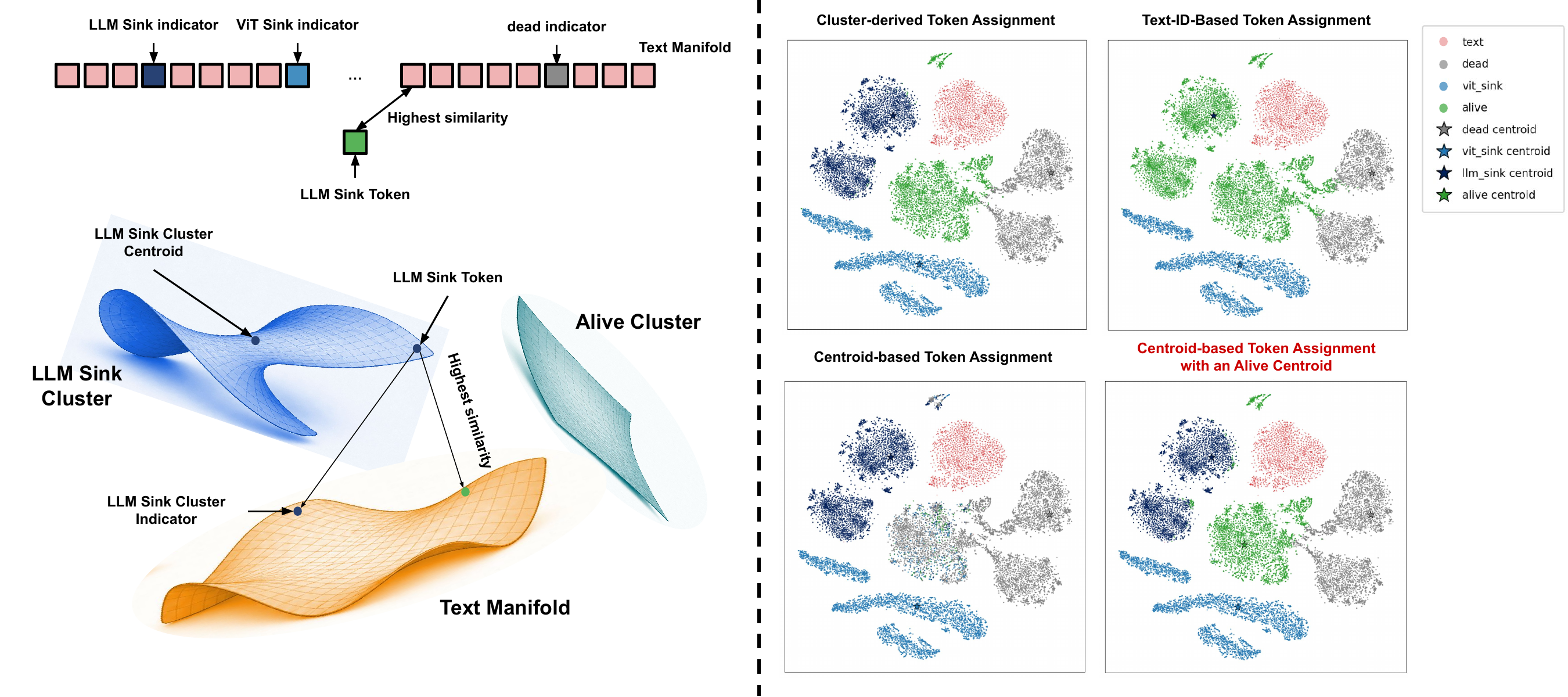}
    \caption{
    Visualization of token role assignment strategies in EmbedLens analysis.
    The left panel illustrates the gap between centroid-level text-ID matching and token-level role assignment, while the right panels show t-SNE visualizations of token roles on COCO-10K under different assignment strategies.
    }
    \label{fig:analysis_tsne_acc}
\end{figure}

\subsection{Refining Token Role Assignment in EmbedLens Analysis}\label{sec:method:EmbedLens_enhanced}

We first revisit how visual token roles are operationally assigned in EmbedLens. As described in Section~\ref{sec:method:preliminaries}, EmbedLens was originally introduced as an embedding-space probing tool that retrieves the nearest textual tokens from the LLM input vocabulary embedding space. 
The token roles are then identified through a cluster-wise analysis of projected visual tokens, where dead tokens correspond to the largest highly homogeneous cross-image cluster, while sink tokens are associated with ViT and LLM sink clusters.

In the original analysis, per-image clusters are first obtained in the projected visual embedding space, and their cross-image consistency is examined through the cosine similarity between cluster centroids. 
Highly consistent clusters are summarized as $C_{S_{\mathrm{ViT}}}$, $C_{S_{\mathrm{LLM}}}$, and $C_D$ for ViT sink, LLM sink, and dead-token clusters, respectively. 
EmbedLens is subsequently used to assign a textual reference ID by retrieving its nearest vocabulary token, yielding reference token IDs $t_{S_{\mathrm{ViT}}}$, $t_{S_{\mathrm{LLM}}}$, and $t_D$.
This provides a practical way to assign token roles at inference time. For each projected visual token $\mathbf{v}$, the nearest textual token ID $t = \operatorname{EmbedLens}_1(\mathbf{v})$ is first retrieved. 
If this ID matches one of the reference IDs associated with non-alive token groups, the visual token is then assigned to the corresponding non-alive category; otherwise, it is treated as an alive token. 

However, as illustrated in Fig.~\ref{fig:analysis_tsne_acc} (left), proximity between a cluster centroid and its reference text embedding does not necessarily imply that all cluster members are equally close to the same reference embedding. 
To examine this issue, we follow EmbedLens and construct a randomly sampled COCO-10K subset from COCO~\cite{lin2014microsoft}, and visualize the resulting token representations using t-SNE. 
In Fig.~\ref{fig:analysis_tsne_acc} (right, upper left), the \textbf{cluster-derived token role assignment} is obtained by directly computing $C_{S_{\mathrm{ViT}}}$, $C_{S_{\mathrm{LLM}}}$, and $C_D$, and visualizing the corresponding cluster members with different colors. 
In contrast, the \textbf{text-ID-based token role assignment} shown in Fig.~\ref{fig:analysis_tsne_acc} (right, upper right) predicts visual token categories solely based on the reference token IDs. We observe that this text-ID-based assignment misclassifies a substantial portion of the LLM sink token cluster as alive tokens.

Moreover, the t-SNE visualization reveals that alive tokens also form a distinct cluster. 
To further verify this observation, we compute the average cross-image cosine similarity for each centroid. The similarities are 0.9994 for $C_{S_{\mathrm{ViT}}}$, 0.9905 for $C_{S_{\mathrm{LLM}}}$, 0.9931 for $C_D$, and 0.8572 for the alive-token cluster. 
Although the relatively lower similarity of alive tokens suggests that they contain more image-specific information, they still form a cluster that is clearly separated from the text manifold. 
This result could be understood as a consequence of the modality gap induced during VLM training.

Based on these observations, we consider two variants for token role assignment. The first is \textbf{centroid-based token role assignment}, which compares each visual token directly with the non-alive centroids rather than relying on their reference text IDs. 
Specifically, a visual token is assigned to a non-alive category if its nearest vector corresponds to one of the non-alive centroids. 
This centroid-based procedure is computationally efficient, as it only requires comparisons with three additional centroid vectors on top of the original comparison with the vocabulary embedding matrix. 
As shown in Fig.~\ref{fig:analysis_tsne_acc} (right, bottom left), this approach correctly identifies LLM sink tokens, but misclassifies alive tokens as dead tokens. 
This suggests that the similarity induced by sharing the same visual modality can dominate the semantic proximity to the text manifold that distinguishes alive tokens from dead tokens.

The second variant is \textbf{centroid-based token role assignment with an alive centroid}. 
In addition to the non-alive centroids, we store an alive-token centroid $C_A$ and classify a visual token as alive not only when it is closest to the text manifold, but also when it is closest to $C_A$. 
As shown in Fig.~\ref{fig:analysis_tsne_acc} (right, bottom right), this variant produces token role assignments that are much closer to the cluster-derived role assignments. 
In the following analyses, we therefore adopt this centroid-based assignment with an alive centroid.

\subsection{Probing the Contribution of Token Roles to Downstream Performance}\label{sec:method:token_role_performance}

In this section, we conduct controlled experiments to examine the functional role of each token category in visual token pruning under a limited token budget. 
Specifically, we consider the three pruning methods under a 50\% token budget, excluding tokens of a specified role from the pruning candidate set. 
This allows us to assess how preserving each token role affects downstream task performance.

\input{tables/filtering_performance_table}

As shown in Table~\ref{tab:role_protected_pruning}, preserving alive tokens consistently improves performance by more than 1\% compared to the original pruning methods. 
This suggests that alive tokens indeed contain semantic information that is important for downstream tasks. 
However, we also observe that preserving non-alive tokens does not necessarily lead to performance degradation.

For FastV and DART, preserving sink tokens results in only minor performance changes, which is consistent with our observation in Fig.~\ref{fig:token_dist_original} that these methods already tend to retain sink tokens. 
Interestingly, we observe that preserving dead tokens can lead to slight performance improvements on six of the ten benchmarks. 

In the case of DivPrune, preserving non-alive tokens causes more noticeable performance changes compared to the original method. 
While preserving sink tokens yields mixed effects depending on the task, preserving dead tokens yields a 0.93\% average gain, although its effect varies across tasks.
These results indicate that, despite their sparse and redundant structure and weak association with image content, preserving dead tokens can still affect downstream task performance.

\subsection{How Token Roles Interact under Visual Token Pruning}\label{sec:method:cross_role_dependencies}

\subsubsection{Attention Structure after Role-Specific Token Removal}
To further examine the attention patterns associated with different token roles inside the LLM, we analyze layer-wise attention mass between token groups.
For each input, we first assign every visual token to one of the four roles. Let $\mathcal{I}_r$ denote the set of token indices belonging to role $r$, where $r \in \{\mathrm{dead}, \mathrm{ViT\text{-}sink}, \mathrm{LLM\text{-}sink}, \mathrm{alive}\}$. 
We also consider textual input tokens and answer tokens as additional target groups. For the $\ell$-th LLM layer, let $\mathbf{A}^{(\ell)}$ be the head-averaged self-attention matrix. 
We define the attention mass from role $r$ to role $s$ as
\begin{equation}
\mathrm{Mass}^{(\ell)}(r,s)
=
\frac{1}{|\mathcal{I}_r|}
\sum_{i \in \mathcal{I}_r}
\sum_{j \in \mathcal{I}_s}
\max\!\left(
\mathbf{A}^{(\ell)}_{i,j},
\mathbf{A}^{(\ell)}_{j,i}
\right).
\end{equation}
where diagonal self-pairs with $i=j$ are excluded when $r=s$. 
This metric captures the aggregate attention associated with two token roles, rather than the average interaction strength of an individual token pair.
We compute this metric for all samples across the 10 benchmark datasets used in our evaluation and report the averaged results. 
This allows us to compare how role-wise attention patterns vary across LLM layers and pruning scenarios.

\begin{figure}[t]
    \centering
    \includegraphics[width=\linewidth]{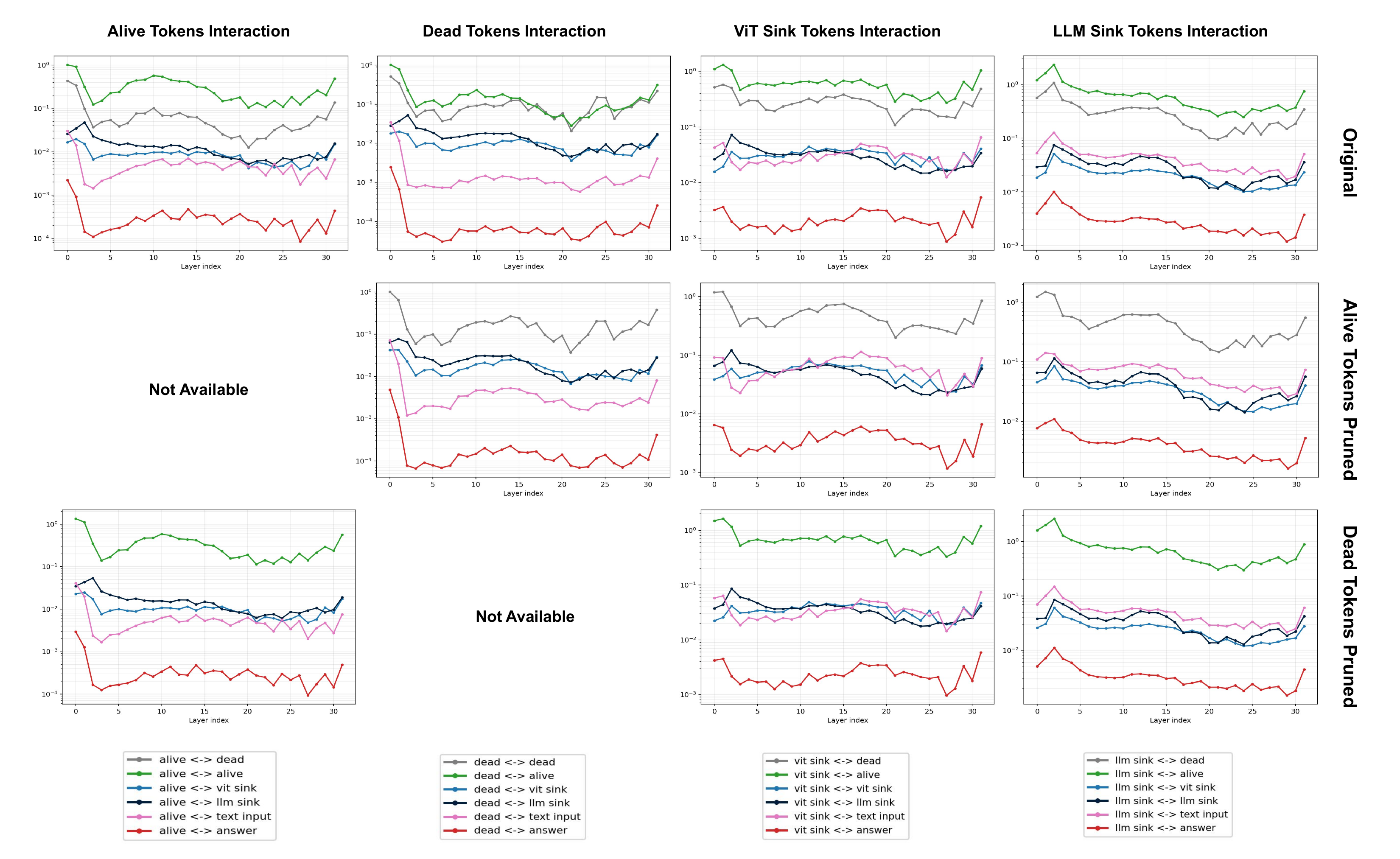}
    \caption{
    Layer-wise group-level attention mass under three input settings:
    the original visual token sequence (top row), alive-token pruning (middle row), and dead-token pruning (bottom row).
    }
    \label{fig:analysis_attn_mass_role_pruning}
\end{figure}

We first analyze the group-level attention patterns under three settings, as shown in Fig.~\ref{fig:analysis_attn_mass_role_pruning}, using the original visual token sequence as input (top row), pruning alive tokens (middle row), and pruning dead tokens (bottom row). 
We summarize two main observations.

\paragraph{Visual–visual token pairs exhibit larger group-level attention mass than visual–textual pairs.}
Across most settings, visual--visual role pairs exhibit larger group-level attention mass than visual--textual pairs.
This trend is observed regardless of the specific token role, suggesting that visual tokens retain strong intra-modal interactions even when they belong to different token categories.

\paragraph{Pruning a specific token role does not appear to substantially alter the remaining attention structure.}
When either alive tokens or dead tokens are pruned, the attention patterns among the remaining token groups remain largely similar to those observed with the original visual sequence. 
This suggests that the broad group-wise attention structure is relatively stable after removing either alive or dead tokens.

\begin{figure}[t]
    \centering
    \includegraphics[width=\linewidth]{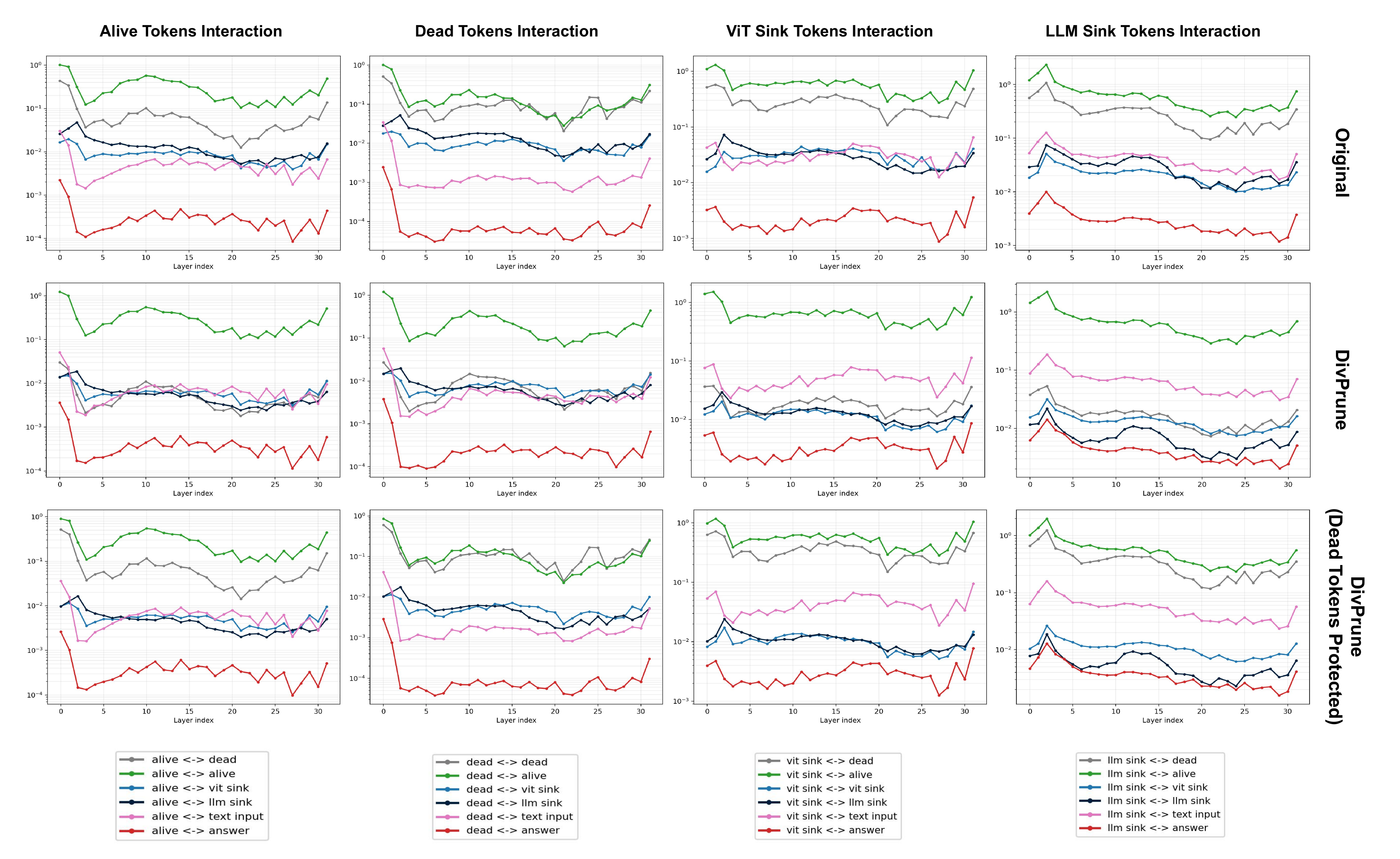}
    \caption{
    Layer-wise group-level attention mass under three input settings:
    the original visual token sequence (top row), DivPrune (middle row), and DivPrune with dead-token protection (bottom row).
    }
    \label{fig:analysis_attn_mass_divprune}
\end{figure}

\subsubsection{Role-wise Attention Changes under DivPrune}

\begin{figure}[t]
    \centering
    \includegraphics[width=\linewidth]{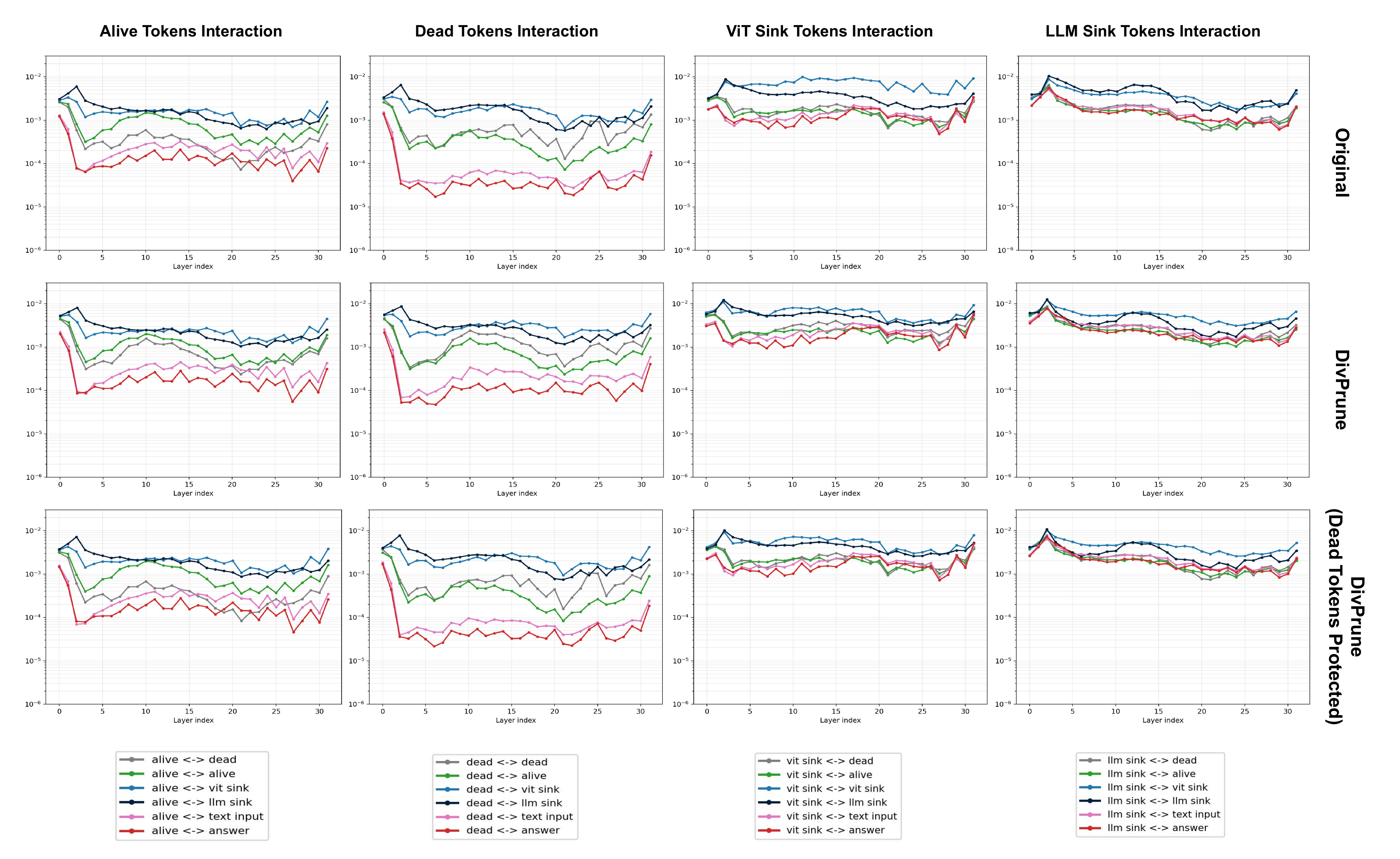}
    \caption{
    Layer-wise pairwise attention score under three input settings:
    the original visual token sequence (top row), DivPrune (middle row), and DivPrune with dead-token protection (bottom row).
    }
    \label{fig:analysis_attn_score_divprune}
\end{figure}

While the preceding analysis removes one token role at a time, practical pruning methods simultaneously change the relative composition of multiple token roles.
Next, we analyze cross-role attention patterns under a practical visual token pruning setting. 
Motivated by the performance improvement observed when protecting dead tokens in DivPrune in Section~\ref{sec:method:token_role_performance}, we focus on DivPrune under a fixed token budget of 50\% and compare the original DivPrune setting with its dead-token-protected variant. 
Since DivPrune tends to prune non-alive tokens more aggressively, as shown in Fig.~\ref{fig:token_dist_original}, attention mass alone does not reveal whether the observed differences are primarily associated with group size or with changes in pairwise interaction strength.
We therefore complement attention mass with a token-level pairwise attention score. 
Specifically, for roles $r$ and $s$ at layer $\ell$, we define
\begin{equation}
\mathrm{Pair}^{(\ell)}(r,s)
=
\frac{1}{|\mathcal{P}_{r,s}|}
\sum_{(i,j) \in \mathcal{P}_{r,s}}
\max\!\left(
\mathbf{A}^{(\ell)}_{i,j},
\mathbf{A}^{(\ell)}_{j,i}
\right),
\end{equation}
where $\mathcal{P}_{r,s}=\mathcal{I}_r \times \mathcal{I}_s$ when $r \neq s$, and $\mathcal{P}_{r,r}=\{(i,j): i,j \in \mathcal{I}_r,\; i \neq j\}$ when $r=s$. 
While attention mass captures the total group-level attention budget allocated between roles, this pairwise score measures the average interaction strength between individual token pairs. 
We report both metrics to analyze whether dead-token protection changes the overall role-level attention allocation or the per-token interaction pattern. 
Fig.~\ref{fig:analysis_attn_mass_divprune} reports the group-level attention mass, while Fig.~\ref{fig:analysis_attn_score_divprune} reports the pairwise attention score. 
Our main observations are summarized as follows.

\paragraph{Dead tokens have low per-token attention but a non-negligible group-level attention mass.}
In Fig.~\ref{fig:analysis_attn_mass_divprune} (top row), dead tokens account for group-level attention mass comparable to that of alive tokens and exhibit a broadly similar layer-wise trend. 
However, Fig.~\ref{fig:analysis_attn_score_divprune} (top row) shows that the average pairwise attention for pairs involving dead tokens remains relatively small. 
This difference is expected because dead tokens form the largest redundant visual-token group. 
Together, these results show that dead tokens collectively account for a non-negligible portion of the group-level visual attention mass.

\paragraph{The reduced non-alive attention mass is largely associated with changes in group sizes.}
Comparing the top and middle rows of Fig.~\ref{fig:analysis_attn_mass_divprune}, we observe that DivPrune reduces the attention mass associated with non-alive tokens, especially dead tokens. 
This is consistent with its tendency to prune a larger fraction of non-alive tokens. 
In Fig.~\ref{fig:analysis_attn_score_divprune}, the pairwise attention scores of the remaining sink tokens change only mildly, suggesting that the reduced sink-token mass is primarily associated with the smaller number of retained sink tokens rather than a large change in their per-token interaction strength. 
For dead tokens, we observe a modest increase in pairwise attention for alive--dead and dead--dead interactions in the middle layers. 
This indicates that the retained dead tokens exhibit slightly higher pairwise interaction scores with alive tokens after pruning.

\paragraph{Dead-token protection appears to yield attention patterns closer to the full-sequence baseline.}
In the dead-token-protected setting, the bottom rows of Fig.~\ref{fig:analysis_attn_mass_divprune} and Fig.~\ref{fig:analysis_attn_score_divprune} show attention patterns involving dead tokens that are closer to those observed with the full visual sequence. 
At the same time, sink tokens are more likely to be pruned, resulting in reduced group-level attention mass for those roles.
Together with the downstream performance changes observed in Section~\ref{sec:method:token_role_performance}, these results suggest that dead tokens should not be treated as uniformly irrelevant under token pruning.
Although individual dead tokens may carry little image-specific semantic information, removing many of them may still affect how the remaining token roles interact when several roles are pruned together.

\subsubsection{Discussion and Implications}

Our results suggest that token roles identified in the projected embedding space do not translate directly into optimal pruning decisions.
While preserving alive tokens consistently improves downstream performance, the effects of preserving non-alive tokens vary across token roles and pruning methods.
In particular, protecting dead tokens can sometimes improve performance, especially for DivPrune, despite their weak association with image-specific semantics at the individual-token level.

The attention analyses provide one possible explanation for this observation.
Although individual dead-token pairs exhibit relatively weak attention, dead tokens collectively account for a non-negligible portion of the group-level attention mass because they form a large and highly redundant token group.
Consequently, aggressively pruning dead tokens can change not only the number of retained tokens, but also the overall composition of token roles and the the attention patterns among the remaining visual tokens.
Importantly, these results should not be interpreted as evidence that dead tokens directly encode useful visual semantics or that they necessarily serve a causal structural function.
Rather, they indicate that the effect of removing a token cannot always be determined solely from its individual semantic relevance.

A practical implication is that visual token pruning may benefit from considering both token-level importance and the composition of the retained token set.
Instead of treating all non-alive tokens as uniformly removable, pruning methods could monitor role-wise retention rates or avoid disproportionately removing a single token category.
Our findings therefore motivate composition-aware pruning as a direction for further investigation, while additional experiments across different VLM backbones, pruning budgets, and token-role assignment strategies are needed to determine how broadly these observations generalize.
\\
\\

%% file: tables/filtering_performance_table.tex
\begin{table*}[t]
\centering
\resizebox{\linewidth}{!}{%
\begin{tabular}{lllllllllllll}
\toprule
Method & Protected Token & MME & POPE & GQA & TextVQA & MMBench & SQA & VQAv2 & VizWiz & OCRBench & SEED & Avg. $\Delta$ (\%) \\
\midrule
\multirow{5}{*}{\textbf{FastV}} & \multicolumn{1}{l}{Original} & 1866.5$_{\textcolor{blue}{0.00\%}}$ & 0.821$_{\textcolor{blue}{0.00\%}}$ & 0.603$_{\textcolor{blue}{0.00\%}}$ & 0.485$_{\textcolor{blue}{0.00\%}}$ & 75.76$_{\textcolor{blue}{0.00\%}}$ & 0.687$_{\textcolor{blue}{0.00\%}}$ & 0.712$_{\textcolor{blue}{0.00\%}}$ & 0.462$_{\textcolor{blue}{0.00\%}}$ & 0.309$_{\textcolor{blue}{0.00\%}}$ & 0.665$_{\textcolor{blue}{0.00\%}}$ & \textcolor{blue}{0.00\%} \\
 & \multicolumn{1}{l}{Alive} & 1863.1$_{\textcolor{blue}{-0.18\%}}$ & \cellcolor{red!10}0.843$_{\textcolor{red}{+2.70\%}}$ & \cellcolor{red!10}0.620$_{\textcolor{red}{+2.85\%}}$ & \cellcolor{red!10}0.494$_{\textcolor{red}{+1.90\%}}$ & \cellcolor{red!10}77.27$_{\textcolor{red}{+2.00\%}}$ & \cellcolor{red!10}0.694$_{\textcolor{red}{+1.01\%}}$ & \cellcolor{red!10}0.719$_{\textcolor{red}{+0.96\%}}$ & 0.454$_{\textcolor{blue}{-1.65\%}}$ & 0.310$_{\textcolor{red}{+0.32\%}}$ & \cellcolor{red!10}0.675$_{\textcolor{red}{+1.52\%}}$ & \cellcolor{red!10}\textcolor{red}{+1.14\%} \\
 & \multicolumn{1}{l}{ViT Sink} & 1866.5$_{\textcolor{blue}{0.00\%}}$ & 0.821$_{\textcolor{blue}{0.00\%}}$ & 0.603$_{\textcolor{blue}{0.00\%}}$ & 0.485$_{\textcolor{blue}{0.00\%}}$ & 75.76$_{\textcolor{blue}{0.00\%}}$ & 0.687$_{\textcolor{blue}{0.00\%}}$ & 0.712$_{\textcolor{blue}{0.00\%}}$ & 0.462$_{\textcolor{blue}{0.00\%}}$ & 0.309$_{\textcolor{blue}{0.00\%}}$ & 0.665$_{\textcolor{blue}{0.00\%}}$ & \textcolor{blue}{0.00\%} \\
 & \multicolumn{1}{l}{LLM Sink} & 1867.8$_{\textcolor{red}{+0.07\%}}$ & 0.821$_{\textcolor{blue}{0.00\%}}$ & 0.603$_{\textcolor{blue}{0.00\%}}$ & 0.483$_{\textcolor{blue}{-0.41\%}}$ & 75.76$_{\textcolor{blue}{0.00\%}}$ & 0.688$_{\textcolor{red}{+0.14\%}}$ & 0.711$_{\textcolor{blue}{-0.11\%}}$ & 0.462$_{\textcolor{blue}{0.00\%}}$ & 0.309$_{\textcolor{blue}{0.00\%}}$ & \cellcolor{red!10}0.669$_{\textcolor{red}{+0.61\%}}$ & \textcolor{red}{+0.03\%} \\
 & \multicolumn{1}{l}{Dead} & 1874.4$_{\textcolor{red}{+0.42\%}}$ & 0.822$_{\textcolor{red}{+0.11\%}}$ & 0.603$_{\textcolor{blue}{-0.01\%}}$ & 0.482$_{\textcolor{blue}{-0.54\%}}$ & \cellcolor{red!10}76.52$_{\textcolor{red}{+1.00\%}}$ & 0.690$_{\textcolor{red}{+0.36\%}}$ & 0.709$_{\textcolor{blue}{-0.45\%}}$ & 0.456$_{\textcolor{blue}{-1.21\%}}$ & \cellcolor{red!10}0.311$_{\textcolor{red}{+0.65\%}}$ & \cellcolor{red!10}0.669$_{\textcolor{red}{+0.61\%}}$ & \textcolor{red}{+0.09\%} \\
\midrule
\multirow{5}{*}{\textbf{DART}} & \multicolumn{1}{l}{Original} & 1870.9$_{\textcolor{blue}{0.00\%}}$ & 0.832$_{\textcolor{blue}{0.00\%}}$ & 0.607$_{\textcolor{blue}{0.00\%}}$ & 0.483$_{\textcolor{blue}{0.00\%}}$ & 78.03$_{\textcolor{blue}{0.00\%}}$ & 0.693$_{\textcolor{blue}{0.00\%}}$ & 0.703$_{\textcolor{blue}{0.00\%}}$ & 0.453$_{\textcolor{blue}{0.00\%}}$ & 0.310$_{\textcolor{blue}{0.00\%}}$ & 0.648$_{\textcolor{blue}{0.00\%}}$ & \textcolor{blue}{0.00\%} \\
 & \multicolumn{1}{l}{Alive} & 1876.5$_{\textcolor{red}{+0.30\%}}$ & \cellcolor{red!10}0.843$_{\textcolor{red}{+1.33\%}}$ & \cellcolor{red!10}0.619$_{\textcolor{red}{+2.00\%}}$ & \cellcolor{red!10}0.495$_{\textcolor{red}{+2.53\%}}$ & 77.27$_{\textcolor{blue}{-0.97\%}}$ & 0.691$_{\textcolor{blue}{-0.36\%}}$ & \cellcolor{red!10}0.720$_{\textcolor{red}{+2.48\%}}$ & \cellcolor{red!10}0.456$_{\textcolor{red}{+0.53\%}}$ & \cellcolor{red!10}0.314$_{\textcolor{red}{+1.29\%}}$ & \cellcolor{red!10}0.671$_{\textcolor{red}{+3.43\%}}$ & \cellcolor{red!10}\textcolor{red}{+1.26\%} \\
 & \multicolumn{1}{l}{ViT Sink} & 1870.9$_{\textcolor{blue}{0.00\%}}$ & 0.832$_{\textcolor{blue}{0.00\%}}$ & 0.607$_{\textcolor{red}{+0.01\%}}$ & 0.483$_{\textcolor{blue}{0.00\%}}$ & 78.03$_{\textcolor{blue}{0.00\%}}$ & 0.693$_{\textcolor{blue}{0.00\%}}$ & 0.703$_{\textcolor{blue}{0.00\%}}$ & 0.453$_{\textcolor{blue}{0.00\%}}$ & 0.310$_{\textcolor{blue}{0.00\%}}$ & 0.648$_{\textcolor{blue}{0.00\%}}$ & \textcolor{blue}{0.00\%} \\
 & \multicolumn{1}{l}{LLM Sink} & 1870.9$_{\textcolor{blue}{0.00\%}}$ & 0.832$_{\textcolor{blue}{0.00\%}}$ & 0.607$_{\textcolor{blue}{0.00\%}}$ & 0.483$_{\textcolor{blue}{0.00\%}}$ & 78.03$_{\textcolor{blue}{0.00\%}}$ & 0.693$_{\textcolor{blue}{0.00\%}}$ & 0.704$_{\textcolor{red}{+0.17\%}}$ & 0.453$_{\textcolor{blue}{0.00\%}}$ & 0.310$_{\textcolor{blue}{0.00\%}}$ & 0.648$_{\textcolor{blue}{0.00\%}}$ & \textcolor{red}{+0.02\%} \\
 & \multicolumn{1}{l}{Dead} & 1873.8$_{\textcolor{red}{+0.15\%}}$ & 0.831$_{\textcolor{blue}{-0.06\%}}$ & 0.607$_{\textcolor{red}{+0.04\%}}$ & 0.474$_{\textcolor{blue}{-1.82\%}}$ & \cellcolor{red!10}78.79$_{\textcolor{red}{+0.97\%}}$ & 0.695$_{\textcolor{red}{+0.29\%}}$ & 0.705$_{\textcolor{red}{+0.31\%}}$ & 0.449$_{\textcolor{blue}{-0.88\%}}$ & 0.309$_{\textcolor{blue}{-0.32\%}}$ & 0.651$_{\textcolor{red}{+0.31\%}}$ & \textcolor{blue}{-0.10\%} \\
\midrule
\multirow{5}{*}{\textbf{DivPrune}} & \multicolumn{1}{l}{Original} & 1797.2$_{\textcolor{blue}{0.00\%}}$ & 0.839$_{\textcolor{blue}{0.00\%}}$ & 0.614$_{\textcolor{blue}{0.00\%}}$ & 0.482$_{\textcolor{blue}{0.00\%}}$ & 74.24$_{\textcolor{blue}{0.00\%}}$ & 0.681$_{\textcolor{blue}{0.00\%}}$ & 0.709$_{\textcolor{blue}{0.00\%}}$ & 0.458$_{\textcolor{blue}{0.00\%}}$ & 0.307$_{\textcolor{blue}{0.00\%}}$ & 0.675$_{\textcolor{blue}{0.00\%}}$ & \textcolor{blue}{0.00\%} \\
 & \multicolumn{1}{l}{Alive} & \cellcolor{red!10}1859.7$_{\textcolor{red}{+3.47\%}}$ & 0.841$_{\textcolor{red}{+0.29\%}}$ & \cellcolor{red!10}0.618$_{\textcolor{red}{+0.62\%}}$ & \cellcolor{red!10}0.498$_{\textcolor{red}{+3.49\%}}$ & 74.24$_{\textcolor{blue}{0.00\%}}$ & \cellcolor{red!10}0.687$_{\textcolor{red}{+0.87\%}}$ & \cellcolor{red!10}0.717$_{\textcolor{red}{+1.07\%}}$ & 0.458$_{\textcolor{blue}{-0.13\%}}$ & \cellcolor{red!10}0.313$_{\textcolor{red}{+1.95\%}}$ & \cellcolor{red!10}0.681$_{\textcolor{red}{+0.90\%}}$ & \cellcolor{red!10}\textcolor{red}{+1.25\%} \\
 & \multicolumn{1}{l}{ViT Sink} & 1792.1$_{\textcolor{blue}{-0.29\%}}$ & 0.838$_{\textcolor{blue}{-0.08\%}}$ & 0.616$_{\textcolor{red}{+0.21\%}}$ & 0.483$_{\textcolor{red}{+0.21\%}}$ & 74.24$_{\textcolor{blue}{0.00\%}}$ & 0.682$_{\textcolor{red}{+0.07\%}}$ & 0.713$_{\textcolor{red}{+0.45\%}}$ & 0.452$_{\textcolor{blue}{-1.44\%}}$ & 0.304$_{\textcolor{blue}{-0.98\%}}$ & 0.673$_{\textcolor{blue}{-0.30\%}}$ & \textcolor{blue}{-0.21\%} \\
 & \multicolumn{1}{l}{LLM Sink} & 1799.9$_{\textcolor{red}{+0.15\%}}$ & 0.839$_{\textcolor{red}{+0.05\%}}$ & 0.616$_{\textcolor{red}{+0.26\%}}$ & 0.480$_{\textcolor{blue}{-0.25\%}}$ & \cellcolor{red!10}75.00$_{\textcolor{red}{+1.02\%}}$ & 0.683$_{\textcolor{red}{+0.22\%}}$ & 0.711$_{\textcolor{red}{+0.20\%}}$ & \cellcolor{red!10}0.462$_{\textcolor{red}{+0.70\%}}$ & 0.305$_{\textcolor{blue}{-0.65\%}}$ & 0.673$_{\textcolor{blue}{-0.30\%}}$ & \textcolor{red}{+0.14\%} \\
 & \multicolumn{1}{l}{Dead} & \cellcolor{red!10}1820.6$_{\textcolor{red}{+1.30\%}}$ & 0.841$_{\textcolor{red}{+0.22\%}}$ & \cellcolor{red!10}0.617$_{\textcolor{red}{+0.50\%}}$ & 0.481$_{\textcolor{blue}{-0.08\%}}$ & \cellcolor{red!10}78.79$_{\textcolor{red}{+6.12\%}}$ & \cellcolor{red!10}0.693$_{\textcolor{red}{+1.75\%}}$ & 0.706$_{\textcolor{blue}{-0.48\%}}$ & 0.457$_{\textcolor{blue}{-0.39\%}}$ & \cellcolor{red!10}0.309$_{\textcolor{red}{+0.65\%}}$ & 0.673$_{\textcolor{blue}{-0.30\%}}$ & \cellcolor{red!10}\textcolor{red}{+0.93\%} \\
\bottomrule
\end{tabular}%
}
\caption{Subscripts denote the relative percentage change from the original result for each pruning method. Red and blue denote positive and non-positive changes, respectively. Avg. $\Delta$ reports the macro-average relative change across the ten benchmarks.}
\label{tab:role_protected_pruning}
\end{table*}

%% file: sections/5_conc.tex
\section{Conclusion}
In this work, we study visual token pruning in VLMs from the perspective of token roles. We show that existing pruning methods have distinct biases toward alive, sink, and dead tokens, but these biases do not directly correlate with downstream performance. 
This indicates that pruning decisions cannot be fully explained by token role or individual saliency alone.

We further refine token-role assignment and evaluate role-protected pruning under a fixed token budget. 
Our results show that preserving non-alive tokens can sometimes maintain or improve performance, suggesting that these tokens can still contribute to model behavior beyond direct semantic representation. 
Our attention analysis shows that non-alive tokens, especially dead tokens, can still occupy a non-negligible portion of the visual attention budget through their group-level presence and interactions. 
Overall, our findings suggest that effective visual token pruning should consider token roles jointly, rather than treating each role as an independently removable category.